\IfFileExists{V2/galbot.cls}{\def\VTwoRoot{V2/}}{\def\VTwoRoot{}}
\documentclass[twocolumn]{\VTwoRoot galbot}

\crefname{figure}{Figure}{Figures}
\Crefname{figure}{Figure}{Figures}
\crefname{table}{Table}{Tables}
\Crefname{table}{Table}{Tables}
\crefname{section}{Section}{Sections}
\Crefname{section}{Section}{Sections}
\crefname{subsection}{Section}{Sections}
\Crefname{subsection}{Section}{Sections}
\crefname{appendix}{Appendix}{Appendices}
\Crefname{appendix}{Appendix}{Appendices}
\crefname{equation}{Equation}{Equations}
\Crefname{equation}{Equation}{Equations}

\usepackage{wrapfig}
\usepackage{tabularx}
\usepackage{textcomp}
\usepackage{stfloats}
\usepackage{fancyhdr}
\usepackage{needspace}
\usepackage{url}
\usepackage{verbatim}
\usepackage{graphicx}
\usepackage{titlesec}
\usepackage{tocloft}
\usepackage{adjustbox}
\usepackage{multirow}
\usepackage{tikz}
\usepackage{comment}
\usepackage{amsmath,amssymb}
\usepackage{colortbl}
\usepackage{color}
\usepackage{booktabs}
\usepackage{hyperref}
\usepackage{subcaption}
\RequirePackage{xspace}
\makeatletter
\DeclareRobustCommand\onedot{\futurelet\@let@token\@onedot}
\def\@onedot{\ifx\@let@token.\else.\null\fi\xspace}

\makeatother

\usepackage{makecell}

\usepackage{pifont}
\usepackage{bbding}
\usepackage{fontawesome}
\usepackage{xspace}

\usepackage{float}

\newlength\savewidth

\newcolumntype{x}[1]{>{\centering\arraybackslash}p{#1pt}}
\newcolumntype{y}[1]{>{\raggedright\arraybackslash}p{#1pt}}
\newcolumntype{z}[1]{>{\raggedleft\arraybackslash}p{#1pt}}

\renewcommand{\paragraph}[1]{\vspace{1mm}\noindent\textbf{#1}}

\usepackage{xcolor}
\usepackage{array}
\usepackage{bbm}
\usepackage{collcell,xfp}
\usepackage{pgf}
\usepackage[most]{tcolorbox}
\usepackage{csquotes}
\usepackage[noorphans,vskip=1em,leftmargin=1em]{quoting}
\usepackage{enumitem}
\usepackage{forest}
\usepackage{caption}
\usepackage{longtable}
\definecolor{tablehead}{HTML}{E3EFE9}
\definecolor{tablestripe}{HTML}{F4F6F5}
\definecolor{tablesummary}{HTML}{EAF3EE}
\definecolor{tablerule}{HTML}{61776C}
\newcommand{\reporthead}[1]{{\sffamily\bfseries #1}}
\newcommand{\reporttablestyle}{%
  \small
  \renewcommand{\arraystretch}{1.12}%
  \setlength{\extrarowheight}{0.6pt}%
  \setlength{\tabcolsep}{6pt}%
  \setlength{\heavyrulewidth}{0.6pt}%
  \setlength{\lightrulewidth}{0.3pt}%
  \setlength{\cmidrulewidth}{0.3pt}%
  \setlength{\aboverulesep}{0pt}%
  \setlength{\belowrulesep}{0pt}%
  \arrayrulecolor{tablerule}%
}
\AtBeginEnvironment{table}{\reporttablestyle}
\AtBeginEnvironment{table*}{\reporttablestyle}
\AtEndEnvironment{table}{\par\vspace{8pt}}
\AtEndEnvironment{table*}{\par\vspace{8pt}}
\usepackage[T1]{fontenc}

\renewcommand{\paragraph}[1]{\vspace{1.25mm}\noindent\textbf{#1}}

\usepackage{algorithm}
\usepackage{listings}

\definecolor{codeblue}{rgb}{0.25, 0.5, 0.5}
\definecolor{codekw}{rgb}{0.35, 0.35, 0.75}
\lstdefinestyle{Pytorch}{
    language = Python,
    backgroundcolor = \color{white},
    basicstyle = \fontsize{9pt}{8pt}\selectfont\ttfamily\bfseries,
    columns = fullflexible,
    aboveskip=1pt,
    belowskip=1pt,
    breaklines = true,
    captionpos = b,
    commentstyle = \color{codeblue},
    keywordstyle = \color{codekw},
}

\definecolor{green}{HTML}{009000}
\definecolor{red}{HTML}{ea4335}

\definecolor{linecolor1}{RGB}{246, 248, 239}
\definecolor{linecolor2}{RGB}{230, 234, 217}
\definecolor{linecolor3}{RGB}{211, 222, 190}

\newcommand{\methodname}{PhysRSI-eval\xspace}

\title{Where is Physical RSI capability from?\\ An Observational Scaling Law of LLM-as-policy}

\author[*]{Zekai Li}
\author[*]{Jiayi Yu}
\author[*]{Jinlu Zhang}
\author[*]{Luyang Niu}
\author[*]{Zekun Qi}
\author[]{Jun Zheng}
\author[]{Xuchuan Chen}
\author[]{Dairu Liu}
\author[]{Yintianrun Zhang}
\author[\dagger]{He Wang}
\author[\dagger]{Li Yi}

\affiliation{Galbot Team}

\date{\today}
\galbotdata[Project Page]{https://physrsi-eval.github.io/}

\abstract{
In this paper, we systematically investigate the mechanisms, scaling behavior, and generalization of physical Recursive self-improvement.
We introduce \methodname to quantify physical RSI through a five-axis metric: \textit{First-Attempt Performance, Efficiency, Success, Budget} and \textit{Improvement} from failure. Different VLMs are evaluated on 18 manipulation tasks under a fixed interaction budget.
The analysis of experimental results motivates the \textit{Physical Barrel Hypothesis}: different capabilities may constrain different parts of improvement. Systems with similar public benchmark averages can still differ widely on the five axes, and a system that reaches a high final score may improve late or consume more of its budget. We study observational scaling from two perspectives: predicting the relative performance of VLM agents from their public capabilities, and forecasting later progress from early physical feedback. Together, these analyses ask how quickly and how far an agent can improve.
Moreover, we conduct workspace interventions and then analyze what improvement leaves behind. Removing history reduces late-round success, frozen workspaces remain useful on unseen layouts, and three cross-task cases achieve success earlier. The results suggest that scene-specific adaptation and reusable experience can coexist.
}

\begin{document}
\maketitle
\pagestyle{empty}
\raggedbottom

\fancypagestyle{authornotes}{%
  \fancyhf{}
  \renewcommand{\headrulewidth}{0pt}
  \renewcommand{\footrulewidth}{0pt}
  \fancyfoot[L]{%
    \raisebox{3mm}{\parbox[b]{\textwidth}{%
      \footnotesize\raggedright
      \footnoterule
      $^{*}$ Equal contribution.\quad $^{\dagger}$ Corresponding authors.}}}
}
\thispagestyle{authornotes}

\begin{figure*}[!t]
  \centering
  \vspace{-5pt}
  \includegraphics[width=\textwidth]{\VTwoRoot 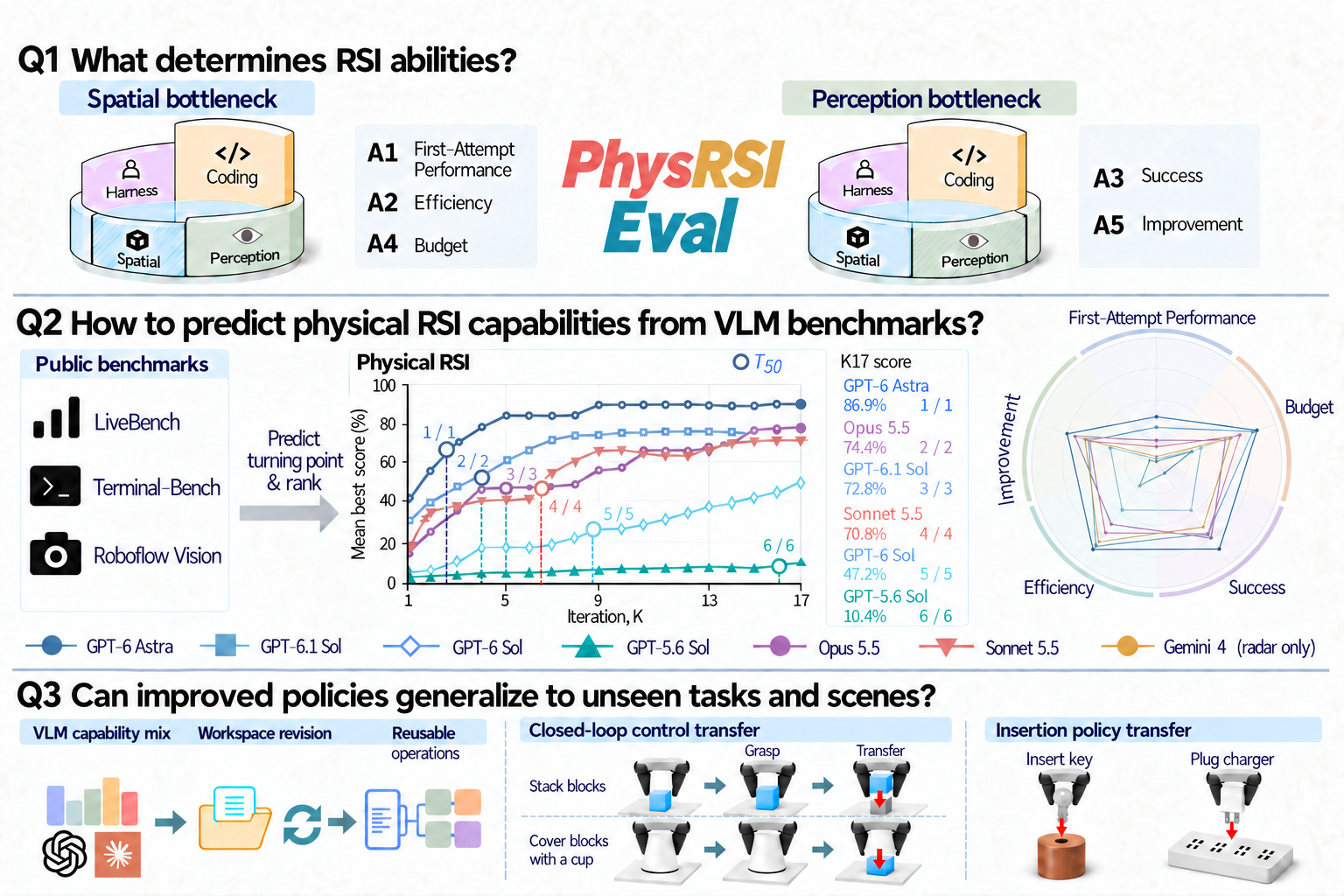}
  \caption{\textbf{\methodname at a glance.} (Q1)~Capability sources and the Physical Barrel Hypothesis. (Q2)~Five-axis quantification and observational scaling of performance, gain timing and model rankings. $T_{50}$ is the half-gain round, the first round at which the best mean score reaches half of its K1--K17 gain. (Q3)~Overfitting and generalization across tasks and scenes.}
  \label{fig:teaser}
\end{figure*}

\section{Introduction}
\label{sec:intro}

Frontier LLM agents demonstrate remarkable capabilities in directly generating actions for robotic manipulation.
Through iterative trial and error, they continually improve manipulation ability by updating skills, generating code, and using tools.~\cite{liang2022codeaspolicies,galbotteam2026astraembodied}
We call this process physical recursive self-improvement, or physical RSI. 
In this work, rather than further demonstrating the impressive capabilities of frontier LLM agents, we pose three more fundamental questions: \textcolor{galbotbg}{\textit{What is the origin of physical RSI capabilities? What scaling laws govern this process? Do the resulting improvements reflect overfitting or generalization?}}

\paragraph{Q1: Where is Physical RSI capability from?}
What allows an agent to turn physical feedback into a better policy? Perception helps it interpret outcomes, spatial reasoning relates observations to actions~\cite{qi2025sofar}, and coding and harness use allow it to implement and test revisions. These capabilities may play different roles. We ask whether strengths behind a good first attempt also explain improvement after feedback.

We introduce \methodname to study these questions under a common protocol (\cref{fig:workflow}). Six systems revise persistent code, tools and notes across 18 manipulation tasks, with frozen model weights and a fixed physical interaction budget. Gemini 4 is reported separately on ten tasks.

An exploratory comparison of four public capability profiles on the 18 tasks associates spatial reasoning most strongly with initial performance, and perception with attained performance and recovery (\cref{fig:barrel}). This pattern motivates the \emph{Physical Barrel Hypothesis}: the capability that limits an agent may depend on which aspect of improvement we measure. Starting well and learning from failure may require different strengths. These associations suggest candidate bottlenecks, not causal requirements.

\Needspace{3\baselineskip}
\paragraph{Q2: How can we quantify Physical RSI and its scaling law?}
An endpoint score alone cannot quantify physical RSI. It says little about when gains occur or what they cost. Repeated attempts can also increase the chance of success without useful revision. Our \emph{five-axis metric} separates initial performance, gain timing, attained score, physical budget savings and recovery from failure (\cref{sec:metrics}). These axes reveal different strengths. Opus 5.5 attains a higher score than GPT-6.1 Sol, but has lower gain-timing and budget-savings scores. Higher scores need not mean improving earlier or using fewer physical interactions.

With these outcomes defined, we study \emph{observational scaling} across model capabilities and physical interaction budgets. Following~\citet{ruan2024observational}, we first ask whether public capability scores can predict the relative order of held-out systems, in attained performance and gain timing (\cref{sec:exp-pca}).

We then ask whether early physical progress predicts the gains from a larger interaction budget. This connects physical RSI to the study of learning curves~\cite{kaplan2020scalinglaws,hoffmann2022chinchilla}, with the interaction round as our budget variable. Calibrating the improvement shape of a reference model to a few early rounds forecasts the rest of a related model's trajectory (\cref{fig:astra-transfer}). The two analyses address complementary questions: what improvement to expect from a model's capabilities, and how much further it may progress after its first attempts. Both remain exploratory.

\paragraph{Q3: Does Physical RSI reflect overfitting or generalization?}
Does the agent acquire a reusable procedure, or adapt to its practice scene? A rising score alone cannot answer this question. We first remove history to test whether retained experience helps beyond repeated attempts. We then freeze workspaces on unseen layouts and transfer them across tasks.

In a six-task Astra ablation, removing history lowers late-round success from 41.7\% to 8.3\% (\cref{tab:app-ablation-rounds}). With further revision disabled, frozen workspace files support success in 25 of 60 trials on unseen layouts (\cref{tab:app-frozen}). Three cross-task cases also reach first success earlier when initialized from other tasks' workspaces (\cref{fig:generalize}). These results support a role for retained experience and show that some artifacts remain useful beyond the scenes where they were refined. Scene-specific gains and useful transfer can coexist. Explaining which retained skills or memories enable transfer connects this question back to the origins of physical RSI.


Our contributions follow this progression:
\begin{itemize}[leftmargin=*, itemsep=1pt, topsep=2pt]
  \item \textbf{Five-axis evaluation and capability diagnosis.} A fixed-budget protocol separates five physical RSI outcomes and relates them to four capability profiles, motivating the Physical Barrel Hypothesis.
  \item \textbf{Observational scaling and prediction.} Cross-model rank prediction and early-round forecasting probe how model capabilities and interaction budget relate to physical improvement.
  \item \textbf{Specialization and generalization.} Ablations, frozen-workspace tests and cross-task adaptation examine whether gains depend on retained experience and whether that experience remains useful elsewhere.
\end{itemize}


\section{Related Work}
\label{sec:related}

\subsection{LLM-as-policy Agents and Embodied Evaluation}
LLMs can act as robot policies by writing programs over perception and control tools~\cite{liang2022codeaspolicies}. Feedback lets them revise plans or reward code~\cite{singh2022progprompt,huang2022innermonologue,ma2023eureka}. Recent benchmarks evaluate these policies through controlled robot interfaces~\cite{fu2026capx,lyu2026codeactionbench,diao2026liberoagent,zhang2026vabench,yang2026robotworld}, while others give coding agents extended simulator access and assess the resulting programs~\cite{shen2026embodiedswe,ma2026rlebench}. These evaluations typically summarize outcomes per episode, across independent trials or within a time budget. A reported Astra case, where a self-written fix succeeded once and failed on repetition, highlights the gap between finding a solution and retaining reliable behavior~\cite{galbotteam2026astraembodied}. \methodname tracks judged physical attempts connected by a persistent workspace, measuring both progress and physical cost under a fixed budget.

\subsection{Agent Self-Improvement}
Agents can accumulate skills and reflections without updating their model weights~\cite{wang2023voyager,shinn2023reflexion,madaan2023selfrefine}. Recent benchmarks examine whether changes to the harness or retained experience improve later performance~\cite{huang2026evobench,wu2026harnessdev,singh2026agentplasticity,zheng2026seagym}. Producing an update and benefiting from it are distinct questions~\cite{lin2026harnessupdating}. Related designs test memory, transfer and gains relative to cost~\cite{shi2026s3gym,xue2026pastbench,gao2026evoagentbench,jiang2026seaeval,ke2026evoharnessbench,wu2026aspire}. AI-R\&D benchmarks address research engineering under time or compute budgets~\cite{wijk2024rebench,chan2024mlebench,starace2025paperbench,rank2026posttrainbench,chi2026ai4aibench}. These designs motivate separating initial ability, improvement and cost, and testing retained experience against repeated attempts. \methodname brings these questions to physical interaction. Native task scores evaluate each attempt, while workspace interventions test what the agent retains. Clearing the workspace tests whether accumulated experience helps; freezing and transferring it test whether its benefits extend to new layouts and tasks.

\subsection{Capability Benchmarks and Observational Scaling}
Public benchmarks provide candidate measures of perception~\cite{fu2024blink,yue2023mmmu,roboflow2026visionevals}, spatial reasoning~\cite{chen2024spatialvlm,yang2024thinkinginspace,du2024embspatial,jia2025omnispatial}, coding and tool use~\cite{jimenez2023swebench,white2024livebench,jain2024livecodebench,merrill2026terminalbench,xie2024osworld,yao2024taubench}, and broader reasoning and work performance~\cite{phan2025hle,tian2024scicode,patwardhan2025gdpval}. Embodied benchmarks and VLA models address planning and physical reasoning~\cite{yang2025embodiedbench,li2024embodiedagentinterface,geminiroboticsteam2025geminirobotics,zhang2025dreamvla,sun2026vlajepa}, but do not track policy revision from graded attempts. Observational scaling relates benchmark-derived capability factors to agent performance across models~\cite{ruan2024observational}. We predict the relative order of physical improvement, in both extent and timing, from public scores. We also forecast later gains from early physical performance. With few closed systems and no training-compute data, these analyses remain observational and exploratory.

\section{PhysRSI-eval: Benchmark Design}
\label{sec:benchmark}
\label{sec:method}
\methodname makes the three questions measurable by recording both physical attempts and the workspace carried between them. The attempt sequence reveals where an agent starts and how it progresses under a fixed budget. Workspace interventions test what agents retain and whether it transfers.

\subsection{What Must the Benchmark Distinguish?}
A strong first attempt and useful revision are different abilities. We therefore score a first-attempt baseline separately and follow subsequent attempts under a fixed budget. Per-round scores show when gains occur; charged physical steps show what they cost. Reaching a high score, reaching it early and using fewer interactions are separate outcomes.

Success alone does not show whether the agent has accumulated useful experience. Code, tools and notes must remain accessible for intervention. Clearing the workspace tests whether retained experience helps; freezing it on unseen layouts tests whether it remains useful; transferring it to new tasks tests whether it can support a different goal. We report these tests separately from the main profiles in~\cref{sec:q3}.

\subsection{A Controlled Loop of Physical Attempts}
\label{sec:protocol}
\begin{figure*}[!tb]
  \centering
  \vspace{-15pt}
  \IfFileExists{\VTwoRoot assets/fig1_workflow_v2.png}{\includegraphics[width=0.85\textwidth]{\VTwoRoot 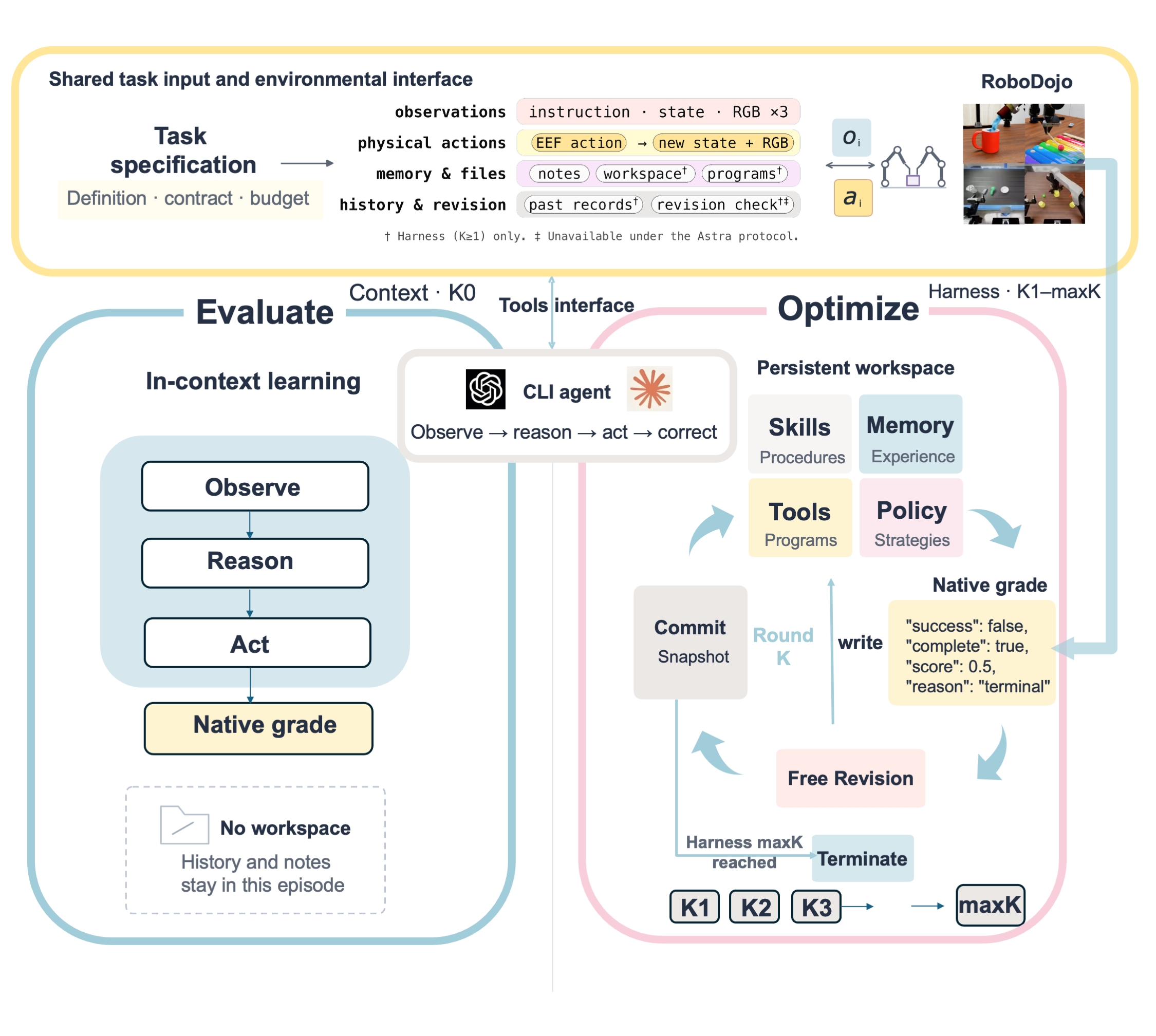}}{\includegraphics[width=0.85\textwidth]{\VTwoRoot 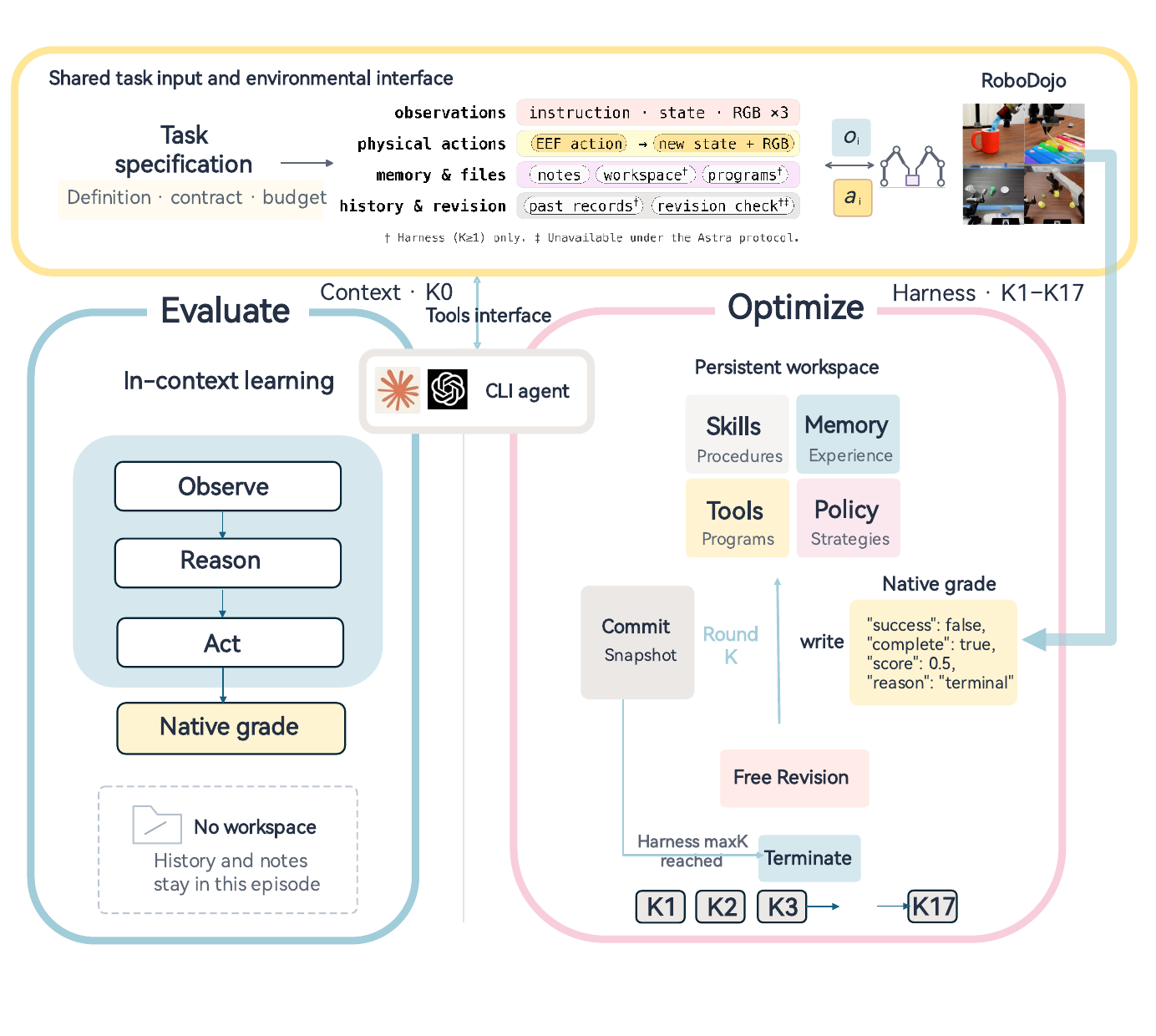}}
  \vspace{-25pt}
  \caption{\textbf{Two stages, one interface.} Top: the task input and robot interface shared by both stages (observations, physical actions, memory and files, history). Left, \emph{Evaluate} (first-attempt round K0): one in-context episode of observe--reason--act through the shared tools, with no workspace; history and notes stay inside the episode. Right, \emph{Optimize} (self-improvement rounds K1--K$H$): each round runs from a persistent workspace of skills, tools, policy and memory, is graded natively, revised with robot actions closed, and committed as the start of the next round, until the first native success or the round limit.}
  \label{fig:workflow}
\end{figure*}

Model weights stay fixed, while skills, tools, policy code and memory can change in a persistent workspace. Each revision is tested through a judged physical attempt (\cref{fig:workflow}). This connects workspace changes to an observed outcome and a physical cost.

K0 measures the starting point without a persistent workspace. Each task runs three times through the shared robot tools, and its scores are averaged. At K1, the agent receives a Harness workspace that is empty except for the host's read-only tools. Each round executes one physical attempt and returns the native score, success flag and failure reason. The agent then revises its workspace with robot actions disabled and commits it for the next round. The main evaluation allows at most $H=17$ refinement rounds per task and stops at the first native success.

Both stages share the robot interface. K1 also introduces workspace tools and program execution, so the K0-to-K1 gain mixes interface changes with improvement. We therefore study the Harness rounds as the self-improvement segment. Each round allows 4200\,s of reasoning; revision takes up to 1800\,s for GPT systems and 600\,s for Claude systems (\cref{tab:app-config}).

For aggregation, successful tasks retain score~1 after stopping, without further execution. Unsuccessful tasks retain their current scores, including regressions. Extra attempts enter the physical budget but add no scoring rounds. Nine GPT-5.6 Sol Harness entries summarize two observed runs. The three-repeat K0 protocol applies to all six systems.

\subsection{Tasks across Difficulty Levels}
\label{sec:experiments}
\begin{figure*}[t]
  \centering
  \includegraphics[width=1.0\textwidth]{\VTwoRoot 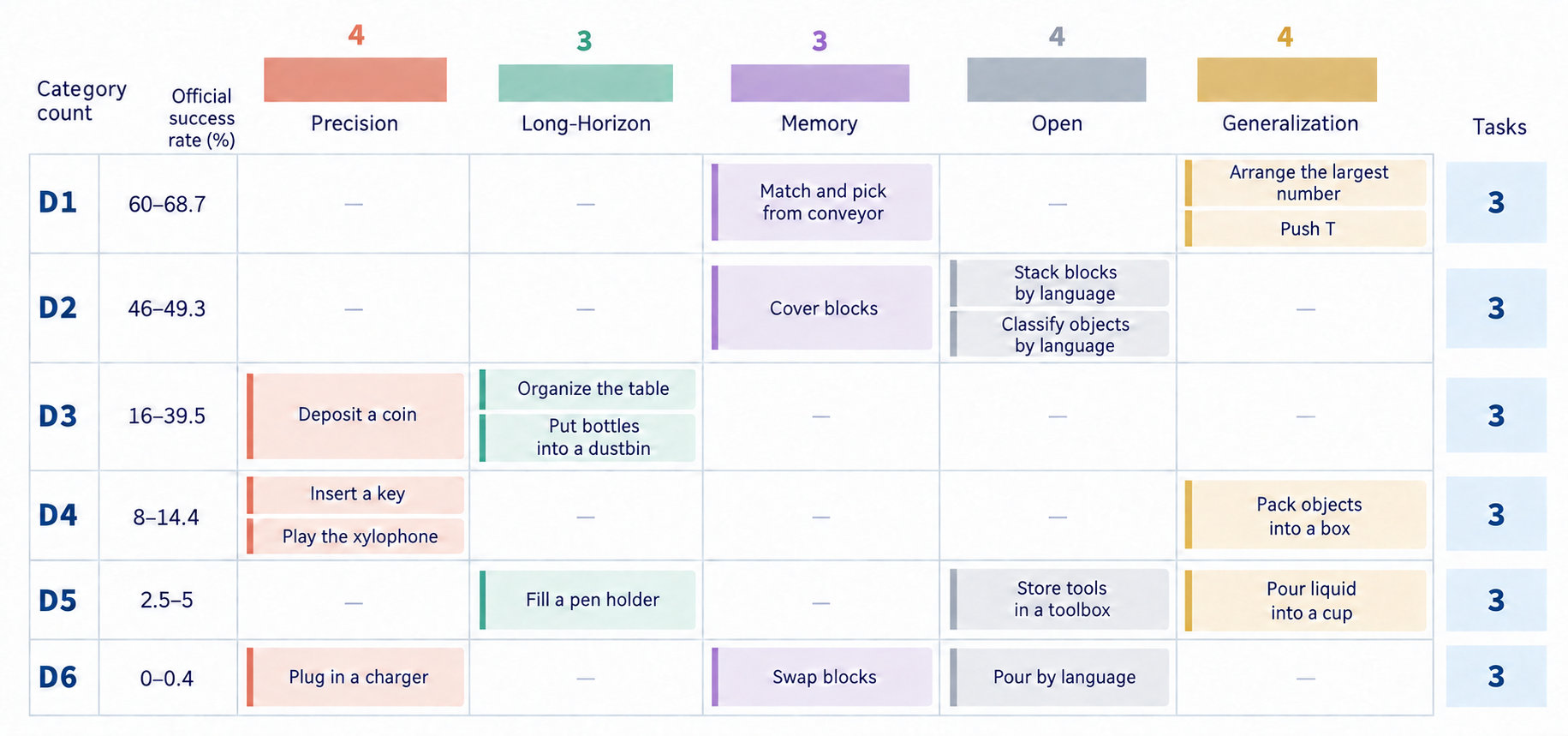}
  \caption{\textbf{Difficulty-balanced task sampling.} Three tasks per difficulty stratum, from D1 (highest official RoboDojo success rate) to D6 (lowest), covering RoboDojo's five official task categories. Strata are assigned by ranking the tasks on the official RoboDojo success rate~\cite{chen2026robodojounifiedsimandrealbenchmark} and are fixed before the main evaluation; they do not describe the subsequent refinement outcomes.}
  \label{fig:tasks}
\end{figure*}

The task suite must expose both immediate competence and room for improvement. We evaluate the loop on 18 RoboDojo dual-arm manipulation tasks. They cover language-conditioned arrangement, stacking, container placement, insertion, pouring, pushing and dynamic picking (\cref{fig:tasks,tab:app-tasks}). The suite spans RoboDojo's five official task categories (Precision, Long-Horizon, Memory, Open and Generalization) and samples three tasks from each of six difficulty strata. The strata are assigned from the official RoboDojo success rates~\cite{chen2026robodojounifiedsimandrealbenchmark}: the 18 tasks are ranked by success rate and grouped three per stratum, from D1 (success rate 60--68.7\%) to D6 (success rate 0--0.4\%). This balances coverage across difficulty levels. The strata are fixed before the main evaluation and are not derived from our refinement outcomes. All 18 tasks, including object packing, receive equal weight. Each retains its native score in $[0,1]$, success criterion and step limit of 300--1300 steps at 25\,Hz. The main evaluation uses layout~0 and seed~0. Partial scores expose progress before the first success.

We evaluate six systems: GPT-6 Astra, GPT-6.1 Sol, GPT-6 Sol, GPT-5.6 Sol, Claude Opus 5.5 and Claude Sonnet 5.5. Each system comprises a model and its runtime configuration, which can also affect performance (\cref{tab:app-config}). In a separate experiment without supplementary hints, Astra solves 13 tasks, compared with 15 in the main evaluation. Results from these conditions are reported separately.

Gemini 4 is reported in a separate partial run covering ten tasks through K20, with six solved. Its reported mean scores are 0.050 at K0 and 0.48 at K10. Its five-axis profile is computed on these ten tasks (\cref{tab:a1a5-reference,fig:radar}). Because the task set differs from the main 18-task protocol, it is excluded from the benchmark-based association and prediction analyses.

\subsection{A Five-Axis Profile of Improvement}
\label{sec:metrics}
To study how physical RSI scales, we need more than an endpoint score. The five-axis profile separates the starting point from gain timing, attained score, physical budget savings and recovery from failure. Each axis measures a different aspect of improvement.

Let $q_k$ be the task-mean score at round $k$, with $q_0$ computed from three-run K0 means. Let $C_k$ be the fraction of tasks solved in Harness rounds K1--K$k$.

\paragraph{A1: First-Attempt Performance.} How well does the agent perform on its first attempt? $A_1=q_0$ is the mean K0 score, averaged over three runs per task and then across tasks, including partial progress.

\Needspace{3\baselineskip}
\paragraph{A2: Efficiency.} How early do gains appear? The timing of gains relative to the terminal net gain is
\begin{equation}
 A_2=\frac{\frac{1}{H-1}\sum_{k=1}^{H-1}q_k-q_0}{q_H-q_0}.
\end{equation}
Earlier sustained gains give higher values. The axis uses K0 as its reference and does not measure wall-clock speed. It is undefined without positive terminal gain; nonmonotone curves are neither clipped nor replaced by best-so-far scores.

\paragraph{A3: Success.} What score does the agent attain? Let $q_{\max}=\max_{0\leq k\leq H}q_k$ and define
\begin{equation}
 \tau_{90}=\min\{k\geq1:q_k\geq q_0+0.9(q_{\max}-q_0)\}.
\end{equation}
Then $A_3=(H-\tau_{90}+1)^{-1}\sum_{k=\tau_{90}}^Hq_k$: the mean over the high window, with partial credit. It is neither a binary success rate nor an asymptotic ceiling.

\paragraph{A4: Budget.} How much physical budget is saved? For a solved task $t$, let $L_t$ be its per-round step limit, $W_t$ its cumulative canonical steps through first success, $Z_t$ the unused steps charged for failed reasoning-timeout rounds, and $e_t$ its charged extra attempts. Define $U_t=(W_t+Z_t)/L_t+e_t$. For the model's solved-task set $\mathcal S$,
\begin{equation}
 A_4=\frac{1}{|\mathcal S|}\sum_{t\in\mathcal S}\max(0,1-U_t/H).
\end{equation}
Higher values mean more of the physical budget was saved. Unsolved tasks do not enter this conditional mean, and the axis is undefined if $\mathcal S$ is empty. Time, tokens and money are not part of it.

\paragraph{A5: Improvement.} Which early failures are overcome? $A_5=(C_H-C_1)/(1-C_1)$ is the fraction of K1-failed tasks subsequently solved within the budget. It is undefined when all tasks succeed at K1. The conditional task sets in A4 and A5 vary across models.
Each axis is reported as a point estimate over the task suite (\cref{app:metrics}).

\section{Where is Physical RSI capability from?}
\label{sec:q1}
\label{sec:exp-barrel}
What allows physical feedback to become a useful policy revision? Acting, interpreting an outcome and implementing a change may draw on different capabilities. We use the five-axis profile to ask whether the capabilities associated with a strong first attempt also track effective improvement.

\subsection{Which Capabilities Track Physical RSI?}
Strong coding scores alone do not ensure physical improvement. GPT-5.6 Sol exceeds GPT-6 Astra on LiveBench code generation and completion, yet solves one of the 18 tasks, whereas Astra solves 15.

\begin{figure*}[t]
  \centering
  \includegraphics[width=1.0\textwidth]{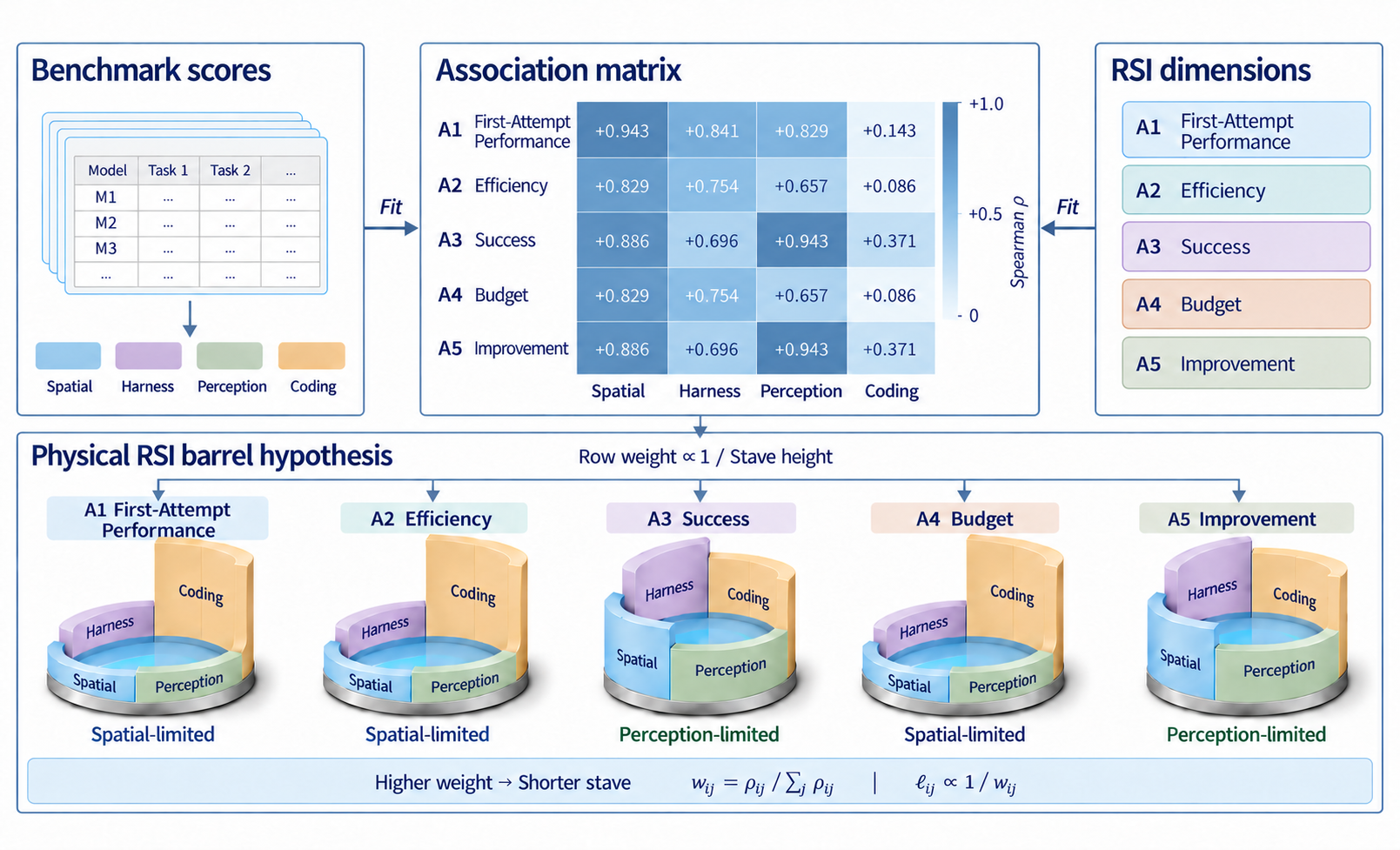}
  \caption{\textbf{Capability associations and the barrel hypothesis.} The matrix reports exploratory rank associations across six systems on the 18 tasks. Each barrel represents one RSI axis, with a stave per capability group. Stave heights illustrate a proposed bottleneck interpretation of the associations.}
  \label{fig:barrel}
\end{figure*}

The exploratory analysis uses six systems on all 18 tasks. We group 17 public benchmark scores into Spatial, Harness, Perception and Coding (\cref{tab:app-benchmarks}), then compare their ranks with A1--A5 (\cref{fig:barrel}).

Spatial reasoning has the strongest observed association with A1 ($\rho=0.94$) and A2/A4 ($\rho=0.83$). Perception has the strongest association with A3 and A5 ($\rho=0.94$). Coding has the weakest association with every axis ($\rho\leq0.37$).

\subsection{The Physical Barrel Hypothesis}
The pattern motivates the \emph{Physical Barrel Hypothesis}: different aspects of improvement may have different capability bottlenecks. Spatial reasoning may matter most for starting well, while perception may matter more for interpreting failures and recovering from them. Under this hypothesis, strong coding cannot always compensate for a weakness elsewhere. Its weak associations do not imply that coding is unnecessary; all systems use code here.

These associations identify candidate bottlenecks that targeted capability interventions can test directly (\cref{app:prediction}).

These candidate bottlenecks lead to the second question: how can we quantify physical RSI and its scaling with model capabilities and interaction budget?

\section{How can we quantify Physical RSI and its scaling law?}
\label{sec:q2}
How much can an agent improve, when do its main gains occur, and how much interaction does it need? The five-axis profile turns these questions into measurable outcomes. We first quantify differences among agents, then study how these outcomes scale with model capabilities and physical interaction budget.

\subsection{High Scores, Fast Gains and Low Cost Can Diverge}
\label{sec:exp-axes}
\begin{figure*}[!tb]
  \centering
  \vspace{-10pt}
  \begin{minipage}[c]{0.35\textwidth}\centering\includegraphics[width=\linewidth]{\VTwoRoot 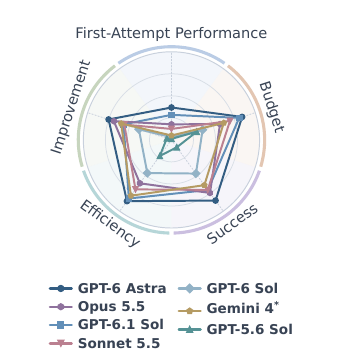}\end{minipage}\hfill
  \begin{minipage}[c]{0.60\textwidth}\centering\includegraphics[width=\linewidth]{\VTwoRoot 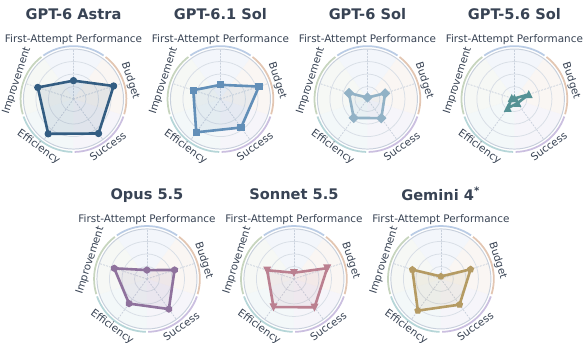}\end{minipage}
  \caption{\textbf{Five-axis profiles.} Left: seven overlaid profiles. Right: one panel per system. Axes clockwise from the top are First-Attempt Performance (A1), Budget (A4), Success (A3), Efficiency (A2) and Improvement (A5), on a shared 0--1 scale. The six reference systems use all 18 tasks, as in~\cref{tab:a1a5-reference}; $^{*}$Gemini 4 uses its ten evaluated tasks (\cref{sec:experiments}).}
  \label{fig:radar}
\end{figure*}

Opus has a higher Success score than GPT-6.1 Sol (0.744 versus 0.703), but lower Efficiency (0.609 versus 0.822). On each system's solved tasks, Opus saves less of the budget (0.579 versus 0.808), and the gap persists on the ten tasks both systems solve. Higher attained scores can come with later gains and greater physical costs (\cref{tab:a1a5-reference,fig:radar}).

Across the six systems, task-mean scores rise from 0.01--0.37 at K0 to 0.10--0.84 by the end of evaluation. They solve 15, 12, 7, 1, 13 and 11 of 18 tasks, respectively: Astra, GPT-6.1 Sol, GPT-6 Sol, GPT-5.6 Sol, Opus and Sonnet. Astra leads on all five axes. The two Claude systems pair relatively high attained scores with middling gain timing, while the GPT Sol family spans a much wider range. GPT-5.6 Sol remains low on Success and Improvement; its Budget score is conditional on one solved task. The profile exposes these trade-offs without a single ranking.

\begin{table*}[!tb]
 \centering
 \caption{\textbf{Five-axis profiles.} K0 scores are averaged over three runs per task. A1--A5 denote First-Attempt Performance, Efficiency, Success, Budget and Improvement. $^{*}$Gemini 4 is evaluated on ten tasks with one K0 run per task.}
 \label{tab:a1a5-reference}
 \begin{tabularx}{\textwidth}{l c *{5}{>{\raggedleft\arraybackslash}X}}
  \toprule
  \rowcolor{tablehead}
  \reporthead{System} & \reporthead{Tasks} & \reporthead{A1} & \reporthead{A2} & \reporthead{A3} & \reporthead{A4} & \reporthead{A5} \\
  \midrule
  \rowcolor{white}
  GPT-6 Astra & 18 & 0.3676 & 0.8594 & 0.8509 & 0.8429 & 0.7500 \\
  \rowcolor{tablestripe}
  GPT-6.1 Sol & 18 & 0.2843 & 0.8224 & 0.7028 & 0.8081 & 0.5714 \\
  \rowcolor{white}
  GPT-6 Sol & 18 & 0.0306 & 0.4658 & 0.4722 & 0.3718 & 0.3889 \\
  \rowcolor{tablestripe}
  GPT-5.6 Sol & 18 & 0.0120 & 0.2247 & 0.1037 & 0.3013 & 0.0556 \\
  \rowcolor{white}
  Claude Opus 5.5 & 18 & 0.1769 & 0.6091 & 0.7444 & 0.5794 & 0.6875 \\
  \rowcolor{tablestripe}
  Claude Sonnet 5.5 & 18 & 0.1241 & 0.6946 & 0.7006 & 0.6984 & 0.5625 \\
  \midrule
  \rowcolor{white}
  Gemini 4$^{*}$ & 10 & 0.0500 & 0.7831 & 0.6350 & 0.6241 & 0.6000 \\
  \bottomrule
 \end{tabularx}
\end{table*}

\begin{figure*}[!tb]
  \centering
  \includegraphics[width=\textwidth]{\VTwoRoot 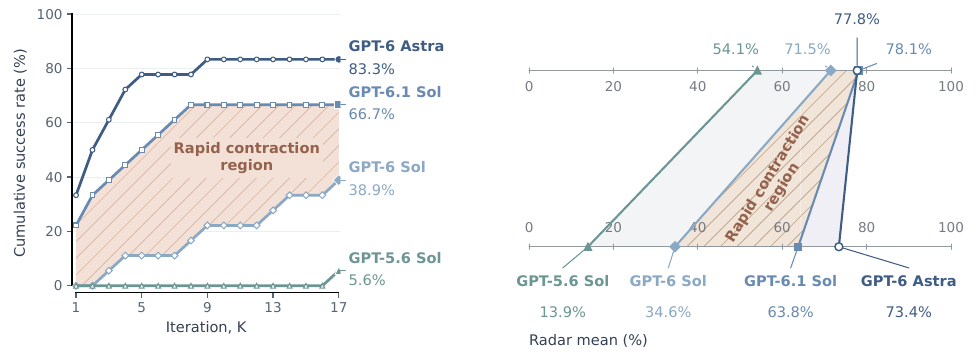}
  \caption{\textbf{Public capability and physical RSI separate differently.} Left: cumulative success during refinement on 18 tasks; the shaded region marks the gap between GPT-6.1 Sol and GPT-6 Sol. Right: mean public score over 17 selected benchmarks versus $100\times(A_1+A_2+A_3+A_4+A_5)/5$. The physical summary is an arithmetic mean, not radar area.}
  \label{fig:killzone}
  \label{fig:benchmark-area}
\end{figure*}

How much do public benchmark averages reveal about these differences? Across the four GPT systems, mean scores over 17 public benchmarks span 54.1--78.1, while the equally weighted physical profile mean, $100\times\sum_{j=1}^{5}A_j/5$, spans 13.9--73.4 (\cref{fig:killzone}). The ranges are 24 and 59.5 points on different scales. GPT-5.6 Sol stays near zero cumulative success for 16 rounds: more attempts do not close every capability gap within this budget.

The Physical Barrel Hypothesis offers one explanation. Useful improvement may require a minimum level of a needed capability, even when other capabilities are strong. The rapid contraction region in~\cref{fig:killzone} illustrates this threshold hypothesis and motivates predicting physical outcomes from capability profiles rather than a public benchmark average.

\subsection{Observational Scaling across Model Capabilities}
\label{sec:exp-pca}
Can public capability scores anticipate physical improvement before the full evaluation is run? This is the cross-model question of observational scaling. Beyond correlating capabilities with outcomes, a predictor must place an unseen system using only the relationships observed in other systems.

\begin{figure*}[!tb]
  \centering
  \includegraphics[width=\textwidth]{\VTwoRoot 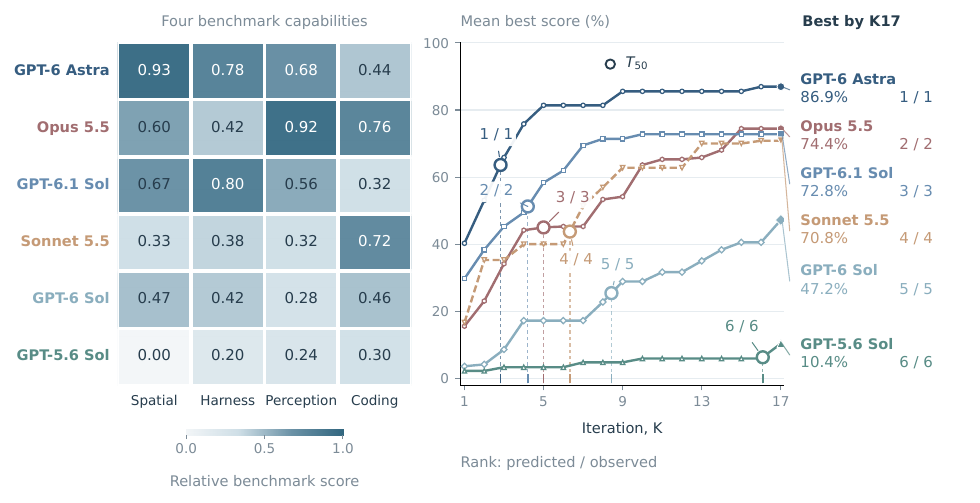}
  \caption{\textbf{Predicting the timing of gains from public capability profiles.} Left: four capability profiles of each system, computed from 17 public benchmarks as normalized ranks among the six systems. Right: best mean score $\bar q_k$ on the 18 tasks. Circles mark $T_{50}$, the first round at which $\bar q_k$ reaches half of its K1--K17 gain. Labels give predicted / observed ranks: $T_{50}$ beside the circles and best score reached by K17 on the right. Each pair of systems is predicted from the other four.}
  \label{fig:k50}
\end{figure*}

We ask how far a system improves and when its gains emerge. Let $\bar q_k=\max_{1\leq j\leq k}q_j$ be the best mean score reached by round $k$. For the first question, ranking systems by the equal-weight average of their four capability profiles reproduces the observed order of $\bar q_{17}$, the best score reached by K17. For the second, we define the half-gain round $T_{50}$ as the first round at which $\bar q_k$ reaches half of its gain from K1 to K17. For every pair of systems, a kernel regressor fitted on the capability profiles of the remaining systems predicts which of the two reaches $T_{50}$ first, and all pairwise orderings are recovered (\cref{fig:k50}). Across the six systems, public capability profiles carry information about both the extent and the timing of physical improvement.

\subsection{What Can Early Physical Feedback Predict?}
Once physical interaction begins, can early progress predict further gains? The six systems reach their high-score windows at different times. The first round reaching 90\% of peak gain, $\tau_{90}$, is K9 for Astra, K7 for GPT-6.1 Sol, K15 for Opus, K13 for Sonnet, and K17 for GPT-6 Sol and GPT-5.6 Sol. Forecasting later rounds tests whether a short trajectory contains useful information about continued improvement.

\begin{figure*}[!tb]
 \centering
 \vspace{-8pt}
 \includegraphics[width=\textwidth]{\VTwoRoot 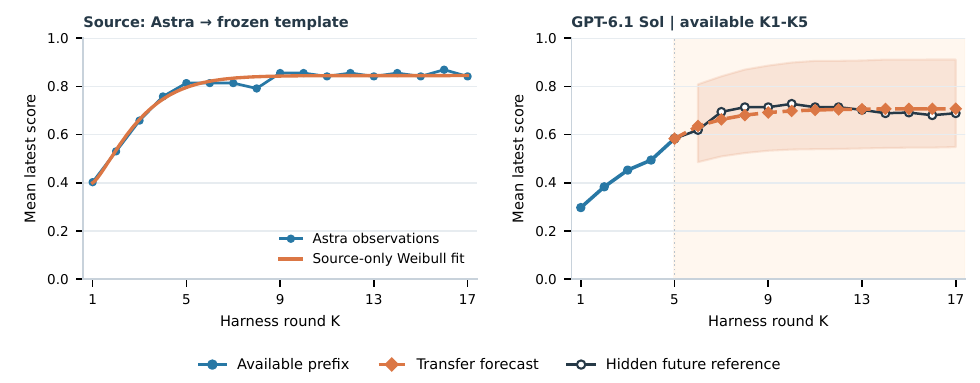}
 \caption{\textbf{Forecasting GPT-6.1 Sol from Astra's improvement shape.} Left: Astra's mean score and the Weibull template fitted to it. Right: GPT-6.1 Sol with K1--K5 available. The template is calibrated on these rounds and forecasts the hidden rounds through K17; shading shows the 90\% paired-task bootstrap band. All 18 tasks are used.}
 \label{fig:astra-transfer}
 \vspace{-5pt}
\end{figure*}

A related model may share the shape of its improvement curve. We fit a Weibull growth template $g_A$ to Astra's K1--K17 mean score and forecast GPT-6.1 Sol as $\hat q(k)=u+(v-u)\,g_A(\rho(k-1))$. The start $u$, plateau $v$ and relative speed $\rho$ are calibrated on Sol's first $K_i$ rounds only, with gain and speed regularized toward the template (\cref{fig:astra-transfer}). With K1--K5 observed, the transferred curve forecasts the remaining rounds through K17 with mean absolute error 0.019, compared with 0.112 for holding the last observed score. A reference trajectory can predict a related model's later performance from fewer rounds.

These analyses address the amount and timing of improvement. We next ask what the gains represent: adaptation to the refinement setting, or experience that remains useful elsewhere?

\section{Does Physical RSI reflect overfitting or generalization?}
\label{sec:q3}
\label{sec:exp-generalization}
With fixed model weights, generalization concerns the code, tools and notes the agent retains. Do they encode reusable procedures, or do they become specialized to the refinement setting? We first test whether retained experience helps beyond repeated attempts. We then evaluate frozen workspaces on unseen layouts and transfer them to new tasks.

\begin{figure*}[!tb]
  \centering
  \vspace{-12pt}
  \includegraphics[width=0.93\textwidth]{\VTwoRoot 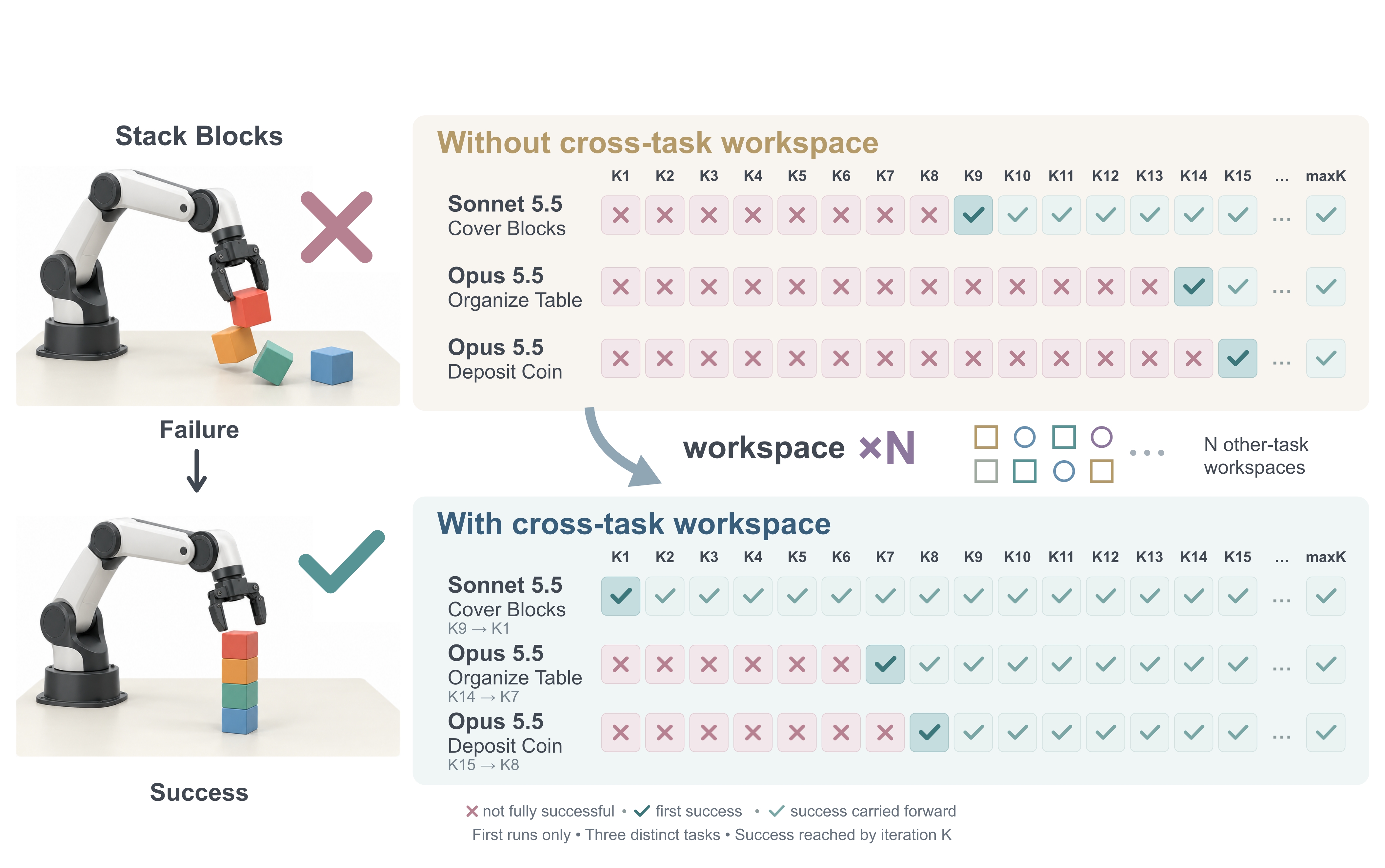}
  \caption{\textbf{Cross-task workspace transfer.} Top: original self-improvement trajectories for tasks first solved at K6 or later. Bottom: trajectories initialized from a workspace pooled from all other tasks, excluding the target task. First success advances from K9 to K1 for Sonnet 5.5 on Cover Blocks, from K14 to K7 for Opus 5.5 on Organize Table, and from K15 to K8 for Opus 5.5 on Deposit Coin. These examples illustrate faster adaptation with transferred workspaces. Red crosses indicate unsuccessful attempts, dark checks mark first success, and pale checks carry success forward without additional evaluation.}
  \label{fig:generalize}
\end{figure*}


\subsection{Does Retained Experience Help Beyond Repeated Attempts?}
\label{sec:exp-ablation}
The Astra ablation varies which information from earlier rounds reaches the agent: the workspace, grader feedback, images and action records (\cref{tab:app-ablation-channels}). It covers six tasks with two chains each; layouts change every round over K1--K6, and execution continues after success. The full loop keeps all four channels, whereas the no-history condition keeps none, so its six rounds are independent attempts. Late-round success over K4--K6 is 15/36 (41.7\%) for the full loop and 3/36 (8.3\%) without history (\cref{tab:app-ablation-rounds}). This supports a role for accumulated information under this protocol.

The remaining conditions withhold individual channels. Withholding grader feedback causes the largest drop, to 5/36 (13.9\%), and withholding earlier-round images yields 12/36 (33.3\%). Keeping only the workspace and grader feedback yields 17/36 (47.2\%), so past images and action records add little once these two channels are kept. The no-history control supports a role for retained experience, but leaves open whether that experience is specific to the refinement setting.

\subsection{What Survives When Revision Stops?}
To test whether the stored artifacts work beyond practice, we freeze Astra workspaces after K6 and test each on five unseen layouts, with no further revision or outcome feedback (\cref{tab:app-frozen}). Across 12 sources, workspace files alone succeed in 25/60 tests (41.7\%). Adding prior raw records gives 23/60 (38.3\%). The stored artifacts remain useful under layout changes, while these point estimates show no benefit from exposing the raw history at test time.

Fixed-layout source workspaces reach 27/60 (45.0\%) on held-out layouts, compared with the reported 14/16 (88\%) on their own layout. The different denominators prevent a matched paired interpretation. Still, the comparison suggests that scene-specific gains and transferable experience can coexist. These tests establish some cross-scene usefulness for Astra on this task subset; they do not establish generalization across all systems.

\subsection{Can Experience Help on a Different Task?}
Traces from five models suggest two forms of reusable experience: executable routines and textual operating knowledge. Sonnet 5.5 and Opus 5.5 increasingly encode control in self-written tools, which generate over 96\% of their actions in later rounds. GPT-6 Sol also builds a substantial tool library, but tools generate 67\% of actions while the model generates 33\% directly. These shares measure tool reliance.

\begin{figure*}[t]
    \centering
    \includegraphics[width=\textwidth]{\VTwoRoot 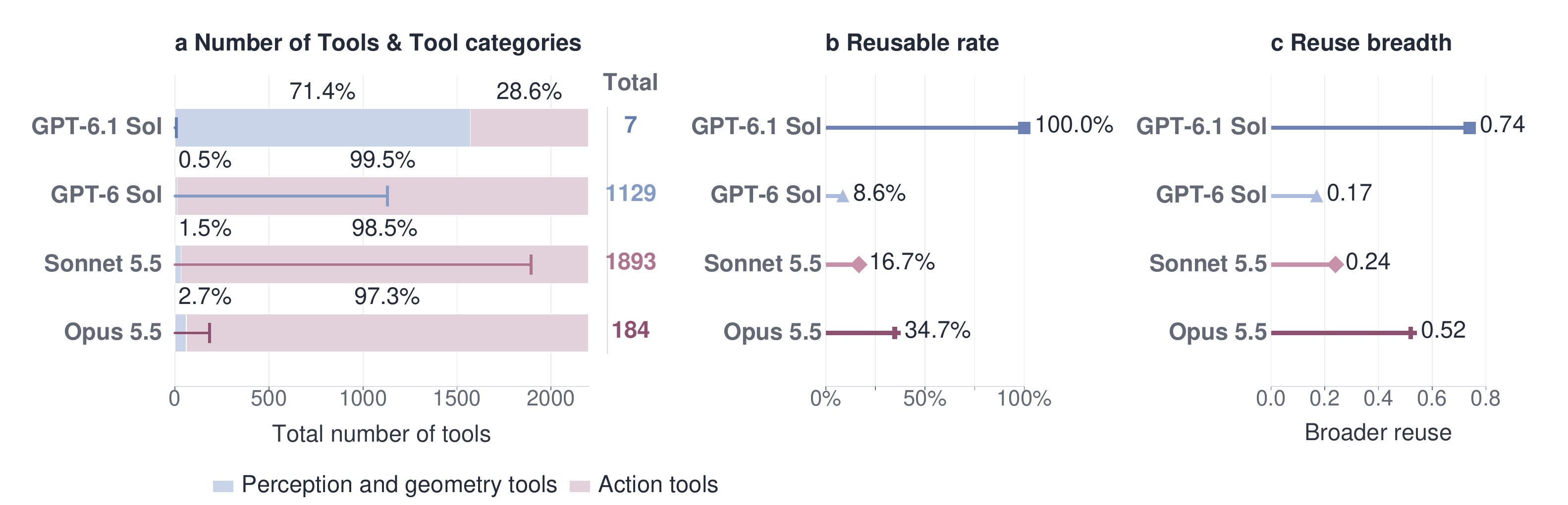}
    \vspace{2pt}
    \caption{\textbf{Tool Libraries and Reuse across Models.} (a) Total tool counts and the proportions of perception and geometry tools versus action tools. (b) Fraction of tool lineages reused in subsequent rounds, among those with later rounds available. Reuse includes direct use, modification and derivation of existing tools. (c) Reuse breadth, defined as one minus the share of subsequent calls accounted for by the top 10\% of reused lineages. Higher values indicate reuse distributed across more lineages. Astra is omitted because it creates very few tools, used only for verification. GPT-6.1 Sol’s estimates reflect a small library of seven tools.}
    \label{fig:tools}
\end{figure*}

These tool libraries grow across rounds, combining tools created from scratch with those derived from earlier implementations. We use tool lineage to identify reuse across rounds. Direct use, modification and derivation of an existing tool all count as reuse. A directed relation A → B indicates that tool B is derived from tool A. Tracing these relations across rounds reveals how later tools build on earlier implementations. Among the three models with substantial tool libraries, Opus exhibits broader reuse and lower concentration. Its lineage reuse rate is 0.347, compared with 0.167 for Sonnet and 0.086 for GPT-6 Sol. The top 10\% of reused lineages account for 48\% of subsequent calls in Opus, compared with 76\% in Sonnet and 83\% in GPT-6 Sol. Together, these patterns suggest that Opus accumulates a broader reusable toolbox, while reuse in Sonnet and GPT-6 Sol remains concentrated in a smaller subset of tools~(\cref{fig:tools}).
Astra and GPT-6.1 Sol follow a different pattern. Approximately 99\% of their actions remain directly generated by the model. They use only a very small number of tools for verification, but their reasoning time decreases across rounds. File-access traces show repeated use of memory, skills and policy notes to recover successful actions and subtask solutions, diagnose failures and guide revisions. These observations suggest that both executable routines and retained operating knowledge can reduce repeated reasoning.

We further find that the workspace and generalization analyses align with the Physical RSI Barrel Hypothesis across its four capability factors: spatial reasoning, perception, coding and harness operation. Spatial tools account for 14.5\% of Sonnet’s tool calls and 13.7\% of Opus’s, with 408 of 2,819 calls and 183 of 1,332 calls, respectively. Astra makes only seven tool calls, including one to such a tool, or 14\%. Astra’s minimal reliance on these modules suggests stronger intrinsic perceptual and spatial capabilities. Sonnet and Opus repeatedly reuse these modules across tasks, suggesting that they rely more on tools to compensate for limitations in these capabilities.

During cross-task transfer, agents also consult memory and skills from semantically similar tasks. In Organize Table, Opus combines a parameterized move tool from Pack Objects Into Box with a memory from Store Tools In Toolbox: “the PITCHED 30 deg grasp... reaches the +y half...” The tool supplies an operation, while the memory identifies a grasp orientation that extends its reach. This shows how a reusable procedure and knowledge of when it works can jointly support transfer~(\cref{fig:generalize}).

The transfer traces suggest that performance gains depend on experience relevant to the target task’s bottleneck and the agent’s ability to assess its applicability from failure evidence. Semantic similarity helps identify candidate experience; failure evidence guides whether to reuse, adapt or discard it. The teaser illustrates this mechanism through the reuse of a closed-loop grasp-and-transfer routine from stacking blocks to covering blocks with a cup, and an insertion routine from key insertion to charger plugging.

\section{Limitations}
\label{sec:limitations}
The main comparison covers six model-plus-configuration systems, with one improvement chain per task. K0 repeats do not estimate variability across complete chains. First-success stopping records one success; reliable repeated execution needs a separate test. Budget and Improvement use conditional task sets that differ across systems. Astra's outcomes differ between the main evaluation and the experiment without supplementary hints.

The capability analysis relies on public benchmark scores whose evaluation settings differ from our runs, and it covers six systems without independent validation. Capability bottlenecks and gain-timing predictions therefore remain hypotheses to test. The ablation and frozen tests cover Astra and six tasks, while cross-task results are three first-run cases. These limits constrain the scope of the observed transfer and the mechanisms that can be inferred from it.

\section{Conclusion}
\label{sec:conclusion}
\methodname studies the origins, measurement, scaling and generalization of physical RSI under a fixed interaction budget. Capability associations motivate the Physical Barrel Hypothesis: the strengths needed to start well may differ from those needed to improve. The five-axis metric makes these outcomes explicit and shows that high scores, early gains and efficient interaction need not coincide. Cross-model prediction and early-round forecasting probe how these outcomes vary with capabilities and interaction budget. Workspace interventions support a role for retained experience and show that it can remain useful on new layouts and tasks. Physical RSI can thus involve both scene-specific adaptation and reusable improvement. Understanding it requires measuring how agents improve and examining what they retain.

\bibliographystyle{\VTwoRoot assets/plainnat}
\bibliography{\VTwoRoot main}

\clearpage
\beginappendix
\crefalias{section}{appendix}
\suppressfloats[t]
\makeatletter
\global\@dbltopnum\z@
\makeatother

\section{Task and Protocol Details}
\label{app:task-protocol}
The 18 tasks keep their native grading and task-specific step limits (\cref{tab:app-tasks}). The main fixed-layout protocol and the changing-layout ablation serve different comparisons and are reported separately.

\paragraph{Actions and observations.} An action sets a target pose for each end-effector, within 5\,cm of translation and 0.35\,rad of rotation of its current pose per call, executed for 1--5 control steps at 25\,Hz; each call returns a state JSON and three camera images. Per round, the reasoning budget is 4200\,s and the revision window after grading is 1800\,s (GPT systems) or 600\,s (Claude systems). The physical budget per task is $17\,L_t$, that is, 17 full-step-limit attempts.

\begin{table}[t]
  \caption{\textbf{The 18 \methodname tasks.} Step limits are per round (25\,Hz). Graded tasks score the highest tier reached (tiers in percent of the native score); binary tasks score 0 or 1. The success check is independent of the tiers.}
  \label{tab:app-tasks}
  \centering
  \setlength{\tabcolsep}{2pt}
  \scriptsize
  \begin{tabular}{y{8} y{96} z{20} y{76}}
    \toprule
    \rowcolor{tablehead}
    \reporthead{\#} & \reporthead{Task} & \reporthead{Steps} & \reporthead{Scoring (tiers, \%)} \\
    \midrule
    \rowcolor{white}
    1 & \texttt{arrange\_largest\_number} & 1050 & graded [5,15,30,100] or [5,15,25,40,100] \\
    \rowcolor{tablestripe}
    2 & \texttt{classify\_objects\_by\_language} & 1100 & graded [10,40,100] \\
    \rowcolor{white}
    3 & \texttt{cover\_blocks} & 800 & graded [0,0,5,15,30,100] \\
    \rowcolor{tablestripe}
    4 & \texttt{deposit\_coin} & 300 & graded [20,100] \\
    \rowcolor{white}
    5 & \texttt{fill\_pen\_holder} & 1100 & graded [10,25,40,90] \\
    \rowcolor{tablestripe}
    6 & \texttt{insert\_key} & 300 & graded [15,100] \\
    \rowcolor{white}
    7 & \texttt{match\_and\_pick\_from\_conveyor} & 700 & binary \\
    \rowcolor{tablestripe}
    8 & \texttt{organize\_table} & 1000 & graded [25,50,75,100] \\
    \rowcolor{white}
    9 & \texttt{pack\_objects\_into\_box} & 1300 & graded [10,25,50,100] \\
    \rowcolor{tablestripe}
    10 & \texttt{play\_Xylophone} & 500 & binary \\
    \rowcolor{white}
    11 & \texttt{plug\_in\_charger} & 400 & binary \\
    \rowcolor{tablestripe}
    12 & \texttt{pour\_by\_language} & 800 & graded [20,50,100] \\
    \rowcolor{white}
    13 & \texttt{pour\_liquid\_into\_cup} & 400 & binary \\
    \rowcolor{tablestripe}
    14 & \texttt{push\_T} & 600 & binary \\
    \rowcolor{white}
    15 & \texttt{put\_bottles\_into\_dustbin} & 700 & graded [10,25,40,100] \\
    \rowcolor{tablestripe}
    16 & \texttt{stack\_blocks\_by\_language} & 400 & graded [20,100] \\
    \rowcolor{white}
    17 & \texttt{store\_tools\_in\_toolbox} & 900 & graded [25,50,75,100] \\
    \rowcolor{tablestripe}
    18 & \texttt{swap\_blocks} & 700 & binary \\
    \bottomrule
  \end{tabular}
\end{table}

\begin{table*}[t]
  \caption{\textbf{System configurations and evaluation coverage.} Each system comprises a model and its runtime configuration, so cross-system comparisons are descriptive. \textsuperscript{a}Settings from session logs; other settings come from configuration files.}
  \label{tab:app-config}
  \centering
  \setlength{\tabcolsep}{4pt}
  \small
  \begin{tabularx}{\textwidth}{y{84} *{6}{>{\raggedright\arraybackslash}X}}
    \toprule
    \rowcolor{tablehead}
    \reporthead{Item} & \reporthead{GPT-6 Astra} & \reporthead{GPT-6.1 Sol} & \reporthead{GPT-6 Sol} & \reporthead{GPT-5.6 Sol} & \reporthead{Opus 5.5} & \reporthead{Sonnet 5.5} \\
    \midrule
    \rowcolor{white}
    Client & Codex CLI 0.156.1 & Codex CLI\textsuperscript{a} & Codex CLI 0.156.1 & Codex CLI 0.156.1 & Claude Code\textsuperscript{a} & Claude Code\textsuperscript{a} \\
    \rowcolor{tablestripe}
    Reasoning effort & xhigh & xhigh\textsuperscript{a} & xhigh & xhigh & high\textsuperscript{a} & xhigh \\
    \rowcolor{white}
    Reasoning budget per round & 4200\,s & 4200\,s\textsuperscript{a} & 4200\,s & 4200\,s & 4200\,s\textsuperscript{a} & 4200\,s \\
    \rowcolor{tablestripe}
    Revision window & 1800\,s & 1800\,s\textsuperscript{a} & 1800\,s & 1800\,s & 600\,s\textsuperscript{a} & 600\,s \\
    \rowcolor{white}
    Selected K1--K17 entries & 91 & 143 & 255 & 306 & 175 & 175 \\
    \rowcolor{tablestripe}
    Counted extra attempts & 13 & 9 & 35 & 12 & 21 & 40 \\
    \rowcolor{white}
    Tasks solved by K17 & 15 & 12 & 7 & 1 & 13 & 11 \\
    \bottomrule
  \end{tabularx}
\end{table*}

\section{Metric Definitions and Computation}
\label{app:metrics}
Metrics are computed after averaging the task curves; in particular, Efficiency is a ratio of aggregate gains, not an average of task-wise ratios. The A3 threshold is the first integer Harness round reaching 90\% of the observed peak gain, without interpolation. Budget charges each extra attempt once, supplements only failed reasoning-timeout rounds to their full step quota, and clips task savings at zero. Undefined denominators stay undefined. The summary in~\cref{fig:killzone} is the equally weighted arithmetic mean of A1--A5, multiplied by 100. It is neither radar polygon area nor a success probability.

\section{Detailed Experimental Results}
\label{app:additional}
The five-axis profiles use the six-system evaluation on 18 tasks; \cref{tab:app-summary} lists the intermediate quantities used to compute them. Astra's scores, curves and costs come from the main evaluation, in which it solves 15 tasks. The separate experiment without supplementary hints solves 13 tasks and is excluded from these aggregates. \Cref{tab:app-first-success} gives the first successful Harness round on each task.

\begin{table}[t]
  \caption{\textbf{Curve summaries for A1--A5 (18 tasks).} $q_0$ and $q_{17}$ are the task-mean scores at K0 and K17, $q_{\max}$ the peak, $\tau_{90}$ the first round reaching 90\% of the peak gain, $C_1$ and $C_{17}$ the fractions of tasks solved by K1 and K17, and ``rec.'' the number of tasks failed at K1 and solved later.}
  \label{tab:app-summary}
  \centering
  \setlength{\tabcolsep}{2pt}
  \scriptsize
  \begin{tabular}{y{46} z{22} z{22} z{22} z{18} z{22} z{22} z{18}}
    \toprule
    \rowcolor{tablehead}
    \reporthead{System} & \reporthead{$q_0$} & \reporthead{$q_{17}$} & \reporthead{$q_{\max}$} & \reporthead{$\tau_{90}$} & \reporthead{$C_1$} & \reporthead{$C_{17}$} & \reporthead{rec.} \\
    \midrule
    \rowcolor{white}
    GPT-6 Astra & 0.368 & 0.842 & 0.869 & 9 & 0.333 & 0.833 & 9 \\
    \rowcolor{tablestripe}
    GPT-6.1 Sol & 0.284 & 0.689 & 0.728 & 7 & 0.222 & 0.667 & 8 \\
    \rowcolor{white}
    GPT-6 Sol & 0.031 & 0.472 & 0.472 & 17 & 0.000 & 0.389 & 7 \\
    \rowcolor{tablestripe}
    GPT-5.6 Sol & 0.012 & 0.104 & 0.104 & 17 & 0.000 & 0.056 & 1 \\
    \rowcolor{white}
    Opus 5.5 & 0.177 & 0.744 & 0.744 & 15 & 0.111 & 0.722 & 11 \\
    \rowcolor{tablestripe}
    Sonnet 5.5 & 0.124 & 0.700 & 0.708 & 13 & 0.111 & 0.611 & 9 \\
    \bottomrule
  \end{tabular}
\end{table}
\begin{table}[t]
  \caption{\textbf{First successful Harness round on each task.} A dash denotes no success within K17.}
  \label{tab:app-first-success}
  \centering
  \setlength{\tabcolsep}{2pt}
  \scriptsize
  \begin{tabularx}{\columnwidth}{>{\raggedright\arraybackslash}X *{6}{x{22}}}
    \toprule
    \rowcolor{tablehead}
    \reporthead{Task} & \reporthead{Astra} & \reporthead{6.1 Sol} & \reporthead{6 Sol} & \reporthead{5.6 Sol} & \reporthead{Opus} & \reporthead{Sonnet} \\
    \midrule
    \rowcolor{white}
    Arrange Largest Number & 1 & 2 & 17 & -- & 3 & 2 \\
    \rowcolor{tablestripe}
    Classify Objects by Language & 1 & 1 & -- & -- & 4 & 2 \\
    \rowcolor{white}
    Cover Blocks & 1 & 3 & 4 & 17 & 10 & 9 \\
    \rowcolor{tablestripe}
    Deposit Coin & 4 & -- & -- & -- & 15 & -- \\
    \rowcolor{white}
    Fill Pen Holder & 3 & 8 & -- & -- & -- & -- \\
    \rowcolor{tablestripe}
    Insert Key & -- & -- & -- & -- & -- & -- \\
    \rowcolor{white}
    Match and Pick from Conveyor & 1 & 1 & 3 & -- & 1 & 1 \\
    \rowcolor{tablestripe}
    Organize Table & 2 & -- & -- & -- & 14 & -- \\
    \rowcolor{white}
    Pack Objects into Box & 9 & 6 & -- & -- & 15 & -- \\
    \rowcolor{tablestripe}
    Play Xylophone & 4 & 4 & 14 & -- & -- & 7 \\
    \rowcolor{white}
    Plug in Charger & 2 & -- & -- & -- & 10 & 7 \\
    \rowcolor{tablestripe}
    Pour by Language & 5 & 7 & 8 & -- & 4 & 13 \\
    \rowcolor{white}
    Pour Liquid into Cup & 2 & 5 & 13 & -- & 3 & 4 \\
    \rowcolor{tablestripe}
    Push T & 3 & 1 & -- & -- & 8 & 8 \\
    \rowcolor{white}
    Put Bottles into Dustbin & 1 & 2 & -- & -- & 1 & 1 \\
    \rowcolor{tablestripe}
    Stack Blocks by Language & 1 & 1 & 9 & -- & 2 & 2 \\
    \rowcolor{white}
    Store Tools in Toolbox & -- & -- & -- & -- & -- & -- \\
    \rowcolor{tablestripe}
    Swap Blocks & -- & -- & -- & -- & -- & -- \\
    \midrule
    \rowcolor{tablesummary}
    \textbf{Solved} & \textbf{15} & \textbf{12} & \textbf{7} & \textbf{1} & \textbf{13} & \textbf{11} \\
    \bottomrule
  \end{tabularx}
\end{table}

\section{Capability and Prediction Analyses}
\label{app:prediction}
The capability profiles use 17 public benchmark scores collected on 15 October 2026 from Roboflow Vision Evals~\cite{roboflow2026visionevals}, ParseBench~\cite{zhang2026parsebench}, LiveBench~\cite{white2024livebench} and five Terminal-Bench 4.0 tasks~\cite{merrill2026terminalbench} (\cref{tab:app-benchmarks}). Each score is converted to a normalized rank among the systems, and ranks are averaged within the Spatial, Harness, Perception and Coding groups. The capability-association matrix compares these group profiles with A1--A5 on the 18 tasks by Spearman correlation, using exact permutation tests over the six systems with Benjamini--Hochberg adjustment; the smallest adjusted $q$ is 0.11. The public-score comparison in~\cref{fig:killzone} averages the 17 raw scores with equal weight. The $T_{50}$ analysis predicts the order of gain timing from the group profiles. For each pair of held-out systems, the group profiles and $T_{50}$ are standardized on the remaining systems, and kernel ridge regression with $k(z,z')=\exp(-\gamma\lVert z-z'\rVert^2)$, $\gamma=0.0625$ and $\alpha=10$ predicts $T_{50}$; the system with the smaller prediction is ranked earlier.

\begin{table}[t]
  \caption{\textbf{Public benchmarks behind the capability profiles.} Scores were collected on 15 October 2026; each group averages normalized benchmark ranks with equal weight.}
  \label{tab:app-benchmarks}
  \centering
  \setlength{\tabcolsep}{3pt}
  \scriptsize
  \begin{tabularx}{\columnwidth}{l >{\raggedright\arraybackslash}X l}
    \toprule
    \rowcolor{tablehead}
    \reporthead{Group} & \reporthead{Benchmark item} & \reporthead{Source} \\
    \midrule
    \rowcolor{white}
    Spatial & Object detection (mAP@50) & Roboflow \\
    \rowcolor{white}
     & OCR text localization & Roboflow \\
    \rowcolor{white}
     & Visual grounding & ParseBench \\
    \rowcolor{tablestripe}
    Harness & ctr-optimization & Terminal-Bench 4.0 \\
    \rowcolor{tablestripe}
     & shadow-relay & Terminal-Bench 4.0 \\
    \rowcolor{tablestripe}
     & Instruction following & LiveBench \\
    \rowcolor{tablestripe}
     & Visual reasoning & Roboflow \\
    \rowcolor{white}
    Perception & OCR transcription & Roboflow \\
    \rowcolor{white}
     & OCR single value & Roboflow \\
    \rowcolor{white}
     & OCR structured JSON & Roboflow \\
    \rowcolor{white}
     & Charts & ParseBench \\
    \rowcolor{white}
     & freecad-platform-drawing & Terminal-Bench 4.0 \\
    \rowcolor{tablestripe}
    Coding & Code generation & LiveBench \\
    \rowcolor{tablestripe}
     & Code completion & LiveBench \\
    \rowcolor{tablestripe}
     & Agentic coding & LiveBench \\
    \rowcolor{tablestripe}
     & mvcc-lsm-compaction & Terminal-Bench 4.0 \\
    \rowcolor{tablestripe}
     & payments-pipeline-fix & Terminal-Bench 4.0 \\
    \bottomrule
  \end{tabularx}
\end{table}

The within-trajectory forecasts use early physical scores rather than public benchmarks. The template $g_A(t)=1-\exp[-(t/\tau_A)^{p_A}]$, with $t=k-1$, is fitted to Astra's K1--K17 mean score and then fixed ($\tau_A=2.18$, $p_A=1.30$). For GPT-6.1 Sol, $u$, $v$ and $\rho$ minimize the squared error on the available rounds plus $\lambda[(v-u)-\Delta_A]^2+(0.03\log\rho/0.2)^2$, where $\Delta_A$ is the gain of the Astra template, $\lambda=4$, $0\leq u\leq v\leq1$ and $\rho$ is searched over 81 log-spaced values in $[0.3,2]$. The remaining rounds are forecast without updating on their outcomes. The band in~\cref{fig:astra-transfer} spans the 5th--95th percentiles of 300 paired task resamples, refitting the template and repeating the calibration in each resample. All 18 tasks are used.

\section{Ablation Details}
\label{app:ablation}
The six tasks are Push T, Plug in Charger, Deposit Coin, Play Xylophone, Pour by Language and Pour Liquid into Cup, with two chains per task and all six rounds run in every chain (no first-success stop). The first chain uses layouts 1--6 and the second layouts 7--12, one per round and in the same order for every condition; the Fixed condition keeps layout 1 or 7 throughout. All rounds are RoboDojo simulation episodes with native grading. Within a round, execution and revision share one conversation, and the next round starts a new conversation. Information from earlier rounds therefore reaches the agent only through the channels in~\cref{tab:app-ablation-channels}. \Cref{tab:app-ablation-rounds} gives the per-round successes. Fixed has 22/36 late-round successes on its seen layouts and is descriptive rather than a matched unseen-layout comparison.

\begin{table}[t]
  \caption{\textbf{Information from earlier rounds in each ablation condition.} Workspace: committed notes, skills and code. Grader: result, score and failure reason after each round, and earlier-round outcome files; step-level success flags in action replies and logs are not part of this channel. Images and actions: observation images and action logs from earlier rounds. Current-round observations, including camera images, are available in every condition.}
  \label{tab:app-ablation-channels}
  \centering
  \scriptsize
  \setlength{\tabcolsep}{2.5pt}
  \begin{tabularx}{\columnwidth}{l >{\raggedright\arraybackslash}X *{4}{c}}
    \toprule
    \rowcolor{tablehead}
    \reporthead{Name} & \reporthead{Condition} & \reporthead{Workspace} & \reporthead{Grader} & \reporthead{Images} & \reporthead{Actions} \\
    \midrule
    \rowcolor{white}
    Full & Full loop & \ding{51} & \ding{51} & \ding{51} & \ding{51} \\
    \rowcolor{tablestripe}
    No-img & No earlier-round images & \ding{51} & \ding{51} & \ding{55} & \ding{51} \\
    \rowcolor{white}
    No-grader & No grader feedback & \ding{51} & \ding{55} & \ding{51} & \ding{51} \\
    \rowcolor{tablestripe}
    WS+FB & Workspace and grader only & \ding{51} & \ding{51} & \ding{55} & \ding{55} \\
    \rowcolor{white}
    No-hist & No history & \ding{55} & \ding{55} & \ding{55} & \ding{55} \\
    \rowcolor{tablestripe}
    Fixed & Full loop, fixed layout & \ding{51} & \ding{51} & \ding{51} & \ding{51} \\
    \bottomrule
  \end{tabularx}
\end{table}

\begin{table}[t]
  \caption{\textbf{Per-round online success in the Astra ablation.} Successes out of 12 chains per condition and round (K1--K6 run in full).}
  \label{tab:app-ablation-rounds}
  \centering
  \scriptsize
  \setlength{\tabcolsep}{2pt}
  \begin{tabular*}{\columnwidth}{@{\extracolsep{\fill}}l *{6}{r}}
    \toprule
    \rowcolor{tablehead}
    \reporthead{$K$} & \reporthead{Full} & \reporthead{Fixed} & \reporthead{No-img} & \reporthead{No-grader} & \reporthead{WS+FB} & \reporthead{No-hist} \\
    \midrule
    \rowcolor{white}
    1 & 1 & 1 & 1 & 0 & 0 & 1 \\
    \rowcolor{tablestripe}
    2 & 2 & 3 & 1 & 2 & 1 & 2 \\
    \rowcolor{white}
    3 & 6 & 6 & 3 & 2 & 2 & 1 \\
    \rowcolor{tablestripe}
    4 & 6 & 8 & 4 & 2 & 7 & 0 \\
    \rowcolor{white}
    5 & 5 & 7 & 3 & 2 & 3 & 0 \\
    \rowcolor{tablestripe}
    6 & 4 & 7 & 5 & 1 & 7 & 3 \\
    \midrule
    \rowcolor{tablesummary}
    \textbf{K4--K6} & \textbf{15/36} & \textbf{22/36} & \textbf{12/36} & \textbf{5/36} & \textbf{17/36} & \textbf{3/36} \\
    \bottomrule
  \end{tabular*}
\end{table}

\section{Generalization Details}
\label{app:generalization}
In the complete frozen-test conditions, each of 12 source chains' K6-revised workspaces is frozen and run once on each of five held-out layouts, with no revision and no outcome feedback (\cref{tab:app-frozen}). Outcomes that share a source or a task are dependent and must not be treated as 60 independent replications.

\begin{table}[t]
  \caption{\textbf{Frozen held-out test of Astra K6 workspaces.} 12 sources $\times$ 5 held-out layouts for the complete conditions; HO-NoImg and HO-NoGrader are incomplete and not comparable with the complete conditions.}
  \label{tab:app-frozen}
  \centering
  \setlength{\tabcolsep}{2.5pt}
  \scriptsize
  \begin{tabular}{y{46} y{68} z{26} z{26} z{24}}
    \toprule
    \rowcolor{tablehead}
    \reporthead{Condition} & \reporthead{Semantics} & \reporthead{Sources} & \reporthead{Succ./$n$} & \reporthead{Rate} \\
    \midrule
    \rowcolor{white}
    E3-full  & Full-loop workspaces; test agent may read prior-round raw records & 12 & 23/60 & 38.3\% \\
    \rowcolor{tablestripe}
    E3-notes & Full-loop workspaces; workspace files only & 12 & 25/60 & 41.7\% \\
    \rowcolor{white}
    HO-Fixed & Fixed workspaces on new layouts & 12 & 27/60 & 45.0\% \\
    \rowcolor{tablestripe}
    HO-NoImg & No-img workspaces; incomplete & 9 (6) & 13/35 & 37.1\% \\
    \rowcolor{white}
    HO-NoGrader & No-grader workspaces; incomplete & 6 (1) & 5/21 & 23.8\% \\
    \bottomrule
  \end{tabular}
\end{table}

\end{document}